\documentclass{article}
\usepackage{spconf,amsmath,graphicx,hyperref}
\usepackage{amssymb,booktabs}
\usepackage{multirow}
\usepackage[percent]{overpic}
\usepackage{xcolor}

\definecolor{inputbox}{HTML}{5B8DB8} 
\definecolor{meshbox}{HTML}{C98242}  

\newcommand{\headvae}{\textsc{PG-VAE}}
\title{ViGA-FLAME: Aligning Latent Visual and Geometric Representations for\\ Robust 3D Head Expression Estimation}

\name{Wen~Hao~He,~Jiacheng~Wu,~Jie Chen$^\dagger$\thanks{$^\dagger$Corresponding Author.}}
\address{Department of Computer Science,\\
Hong Kong Baptist University, Hong Kong SAR}

\begin{document}
\maketitle

\begin{abstract}
Estimating 3D head expression from a single image requires mapping visual appearance to underlying facial geometry. We propose ViGA-FLAME, a two-stage framework that aligns semantic visual features with a structured geometric latent space. First, a Parametric–Geometric Variational Autoencoder (PG-VAE) learns a unified representation of canonical head surfaces and FLAME parameters, supporting both detailed surface reconstruction and interpretable parameter recovery. Supervision on geometry and parameters, together with consistency between the two decoding branches, connects surface detail with parametric structure. Second, a trainable Visual-Geometric Adapter (ViGA) maps features from a pretrained semantic visual encoder into this latent space. Asymmetric attention enables ViGA to align semantic image features from the visual encoder with the structured geometric information represented in the PG-VAE latent space, without altering the original visual stream. By grounding image evidence in this latent prior, ViGA-FLAME enables robust 3D head expression estimation across severe facial deformations and large viewing angles, efficiently improving robustness over competing feed-forward methods.
More visualizations can be found via the \href{https://sapiens2-gpa.github.io/ViGA-FLAME/}{\textit{project page}}.

\end{abstract}
\section{Introduction}

Recovering 3D head geometry and expression from a single image is challenging because facial appearance entangles identity, expression, pose, and viewpoint, while self-occlusion conceals surface deformations. FLAME~\cite{li2017flame} provides a compact, animatable representation of these factors, yet reliably inferring its parameters remains difficult under extreme expressions and profile views. 
Predicting FLAME parameters from images requires visual features that capture appearance cues and a geometric prior for 3D interpretation.

Existing FLAME estimation methods rely on either optimization or feed-forward prediction. 
Optimization-based approaches minimize landmark\cite{bulat2017far, lugaresi2019mediapipe}, photometric~\cite{zielonka2022towards}, and appearance objectives~\cite{qian2024vhap} for each input image. 
Although flexible, these methods incur substantial test-time cost and can be sensitive to initialization, occlusion, and rendering mismatch. Feed-forward methods directly regress parameters or geometry: DECA~\cite{feng2021deca} predicts FLAME parameters and geometric-detail codes, SMIRK~\cite{filntisis2024smirk} and TEASER~\cite{liu2025teaser} leverage appearance-aware supervision for expression recovery, and 
SHeaP~\cite{schoneveld2025sheap} learns a 3D morphable-model predictor through reanimation-based self-supervision with 3DMM-rigged 2D Gaussian rendering.
However, these methods typically regress FLAME parameters directly from image features, without first organizing dense geometry and parametric factors in a shared latent representation.

We propose ViGA-FLAME, a feed-forward framework that maps images into a learned parametric-geometric latent space for 3D head reconstruction.
First, PG-VAE learns a unified parametric-geometric latent space from canonical FLAME surfaces, jointly supporting dense geometric reconstruction with parametric decoding. Second, ViGA augments a frozen image encoder (Sapiens2~\cite{khirodkar2026sapiens2}) with a read-only side-branch adapter, whose semantic and geometric tokens extract visual features without modifying the backbone and project them into the PG-VAE latent space. By aligning visual features with this geometric prior, ViGA-FLAME transfers image context to structured prediction, enabling surface reconstruction and FLAME parameter recovery in a single forward pass.
Our contributions are threefold:
(1) PG-VAE, which unifies dense geometry and parametric FLAME representations;
(2) ViGA, a read-only adapter that aligns visual features with the geometric latent space; and
(3) extensive experiments demonstrating improved accuracy and robustness across shape, expression, pose, and viewpoint variations.

\section{Proposed Method}
\begin{figure*}[htbp]
    \centering
    \includegraphics[width=1\textwidth]{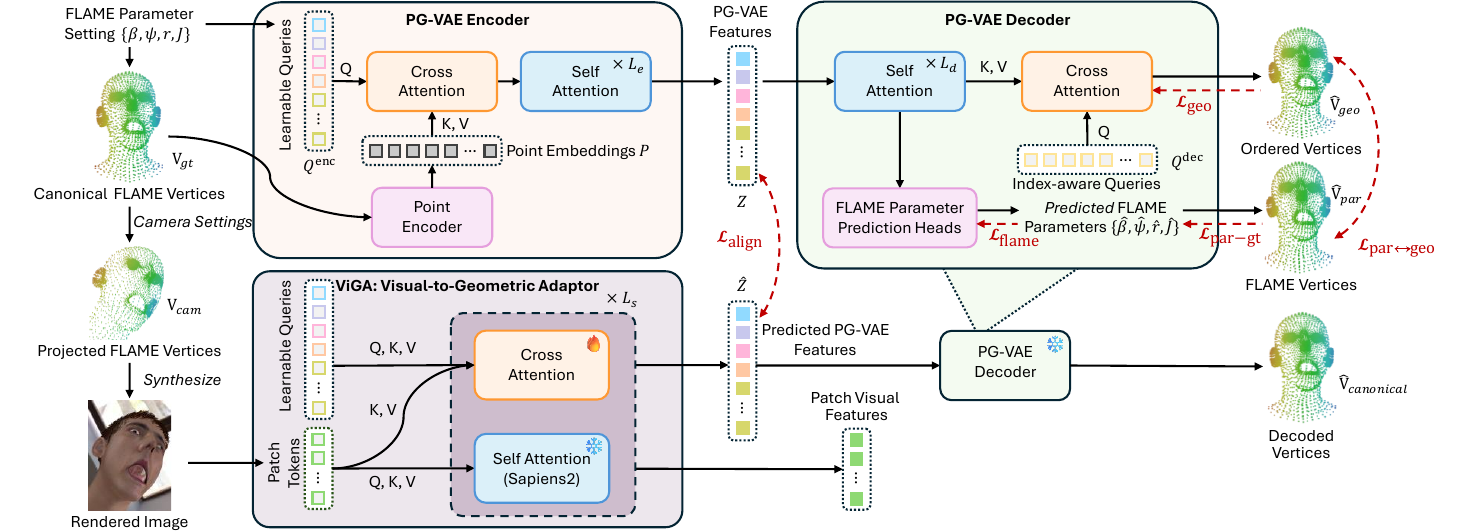}
    \vspace{-0.8cm}
    \caption{\textbf{Method overview}. We first introduce PG-VAE (Sec.~\ref{subsec:headvae}) to learn a unified geometric representation from canonical FLAME surfaces. We then align its frozen representation with image features using the ViGA-FLAME, described in Sec.~\ref{subsec:enhanced_sapiens2}.}
    \label{fig:overview}
    \vspace{-0.2cm}
\end{figure*}

As shown in Fig.~\ref{fig:overview}, our framework first learns a unified parametric-geometric prior from canonical FLAME surfaces using PG-VAE. ViGA then predicts the corresponding latent representation from a single image by aligning visual features with the learned latent space, enabling decoding into either FLAME parameters or dense surface vertices.

\subsection{PG-VAE: Unified Geometry and FLAME Representation}
\label{subsec:headvae}
Parametric head models~\cite{li2017flame} provide interpretable control but constrain geometric expressivity, whereas non-parametric surfaces preserve detail without semantic factorization. PG-VAE aims to combine both by learning a unified latent representation with geometric tokens for dense surface reconstruction and semantic tokens for FLAME parameter recovery.

\textbf{Model Architecture.}
PG-VAE learns a joint parametric-geometric tokenized representation from a canonical, root-centered FLAME mesh $\mathcal{M}=(V,E)$, where $V\in\mathbb{R}^{N\times3}$ contains $N=5023$ vertex coordinates and $E$ denotes the fixed FLAME face connectivity. Fourier features and linear projections encode the vertex coordinates $V$ into point features $P\in\mathbb{R}^{N\times d_e}$, where $d_e$ is the encoder channel size.
The encoder comprises four parameter-conditioned semantic queries and $N_g$ learnable geometric queries. Specifically, the FLAME parameters are denoted by $\{\beta,r,\psi,J\}$, where $\beta$ and $\psi$ represent shape and expression coefficients, $r$ contains the global orientation together with the rotations of the key FLAME joints, and $J$ denotes the corresponding joint positions. Four independent MLPs project these factors into shape, rotation, expression, and position queries, respectively:
\begin{equation}
Q^{enc}=
[q_{\rm shap},q_{\rm rot},q_{\rm exp},q_{\rm pos},
q^{\rm geo}_1,\ldots,q^{\rm geo}_{N_g}]^\top.
\end{equation}
The queries aggregate point features through cross-attention and interact through self-attention, yielding a variational token latent $Z$.
PG-VAE uses geometric and parametric decoders.
The geometric decoder $D_{\mathrm{geo}}$ performs explicit fixed-topology mesh decoding. 
It uses $N$ learnable vertex-indexed queries $Q^{\mathrm{dec}}=[q^{\mathrm{dec}}_1,\ldots,q^{\mathrm{dec}}_N]^\top$, with one query associated with each fixed FLAME vertex index. These queries cross-attend to the decoded latent tokens and regress the corresponding canonical vertex coordinates, yielding $\hat V_{\mathrm{geo}}=D_{\mathrm{geo}}(Z,Q^{\mathrm{dec}})\in\mathbb{R}^{N\times3}$.
This explicit ordered-vertex decoding preserves fixed FLAME vertex correspondence while coupling dense geometric reconstruction with interpretable parametric factors.

\textbf{Training Objectives.}
Let $V_{\mathrm{gt}}$ denote the canonical target head surface, $\hat V_{\mathrm{geo}}$ the output of the geometric decoder, and $\{\hat\beta,\hat r,\hat\psi,\hat J\}$ the predictions of the FLAME-semantic branch.
The predicted FLAME parameters are further decoded into a surface,
$\hat V_{\mathrm{par}}=
\mathcal{D}_{\mathrm{FLAME}}(\hat\beta,\hat r,\hat\psi)$.
We optimize PG-VAE using complementary losses that supervise both decoding branches and explicitly align their geometric outputs.

First, the geometric branch is supervised by a vertex-wise Huber loss $\mathcal{L}_{\mathrm{geo}}$ between $\hat V_{\mathrm{geo}}$ and $V_{\mathrm{gt}}$. Since all meshes share the FLAME topology, this loss directly compares corresponding vertex coordinates and encourages the geometric tokens to preserve fine-grained surface deformations.
Second, $\mathcal{L}_{\mathrm{flame}}$ supervises the FLAME predictions: mean-squared error for shape/expression coefficients, $SO(3)$ geodesic distance for global orientation and joint rotations, and Huber loss for joint positions. This term encourages the semantic tokens to retain explicit control over identity, expression, and articulation.
Third, parameter-space supervision alone is insufficient because equal errors in different FLAME coefficients can induce substantially different surface deformations. We therefore decode the predicted parameters into $\hat V_{\mathrm{par}}$ and impose a vertex-wise Huber loss $\mathcal{L}_{\mathrm{par\text{-}gt}}$ against $V_{\mathrm{gt}}$. This decoded-surface loss grounds parametric estimation in canonical 3D geometry rather than solely in coefficient space.
Fourth, to couple the two branches, we apply a consistency loss $\mathcal{L}_{\mathrm{par\text{-}geo}}$ between $\hat V_{\mathrm{par}}$ and $\hat V_{\mathrm{geo}}$.
Finally, the KL divergence $\mathcal{L}_{\mathrm{KL}}$ regularizes the variational posterior toward a unit Gaussian prior. 
To summarize, the overall objective of PG-VAE is:
\begin{equation}
\mathcal{L}_{\mathrm{PG}}
=
\mathcal{L}_{\mathrm{geo}}
+\lambda_f\mathcal{L}_{\mathrm{flame}}
+\lambda_g\mathcal{L}_{\mathrm{par\text{-}gt}}
+\lambda_c\mathcal{L}_{\mathrm{par\text{-}geo}}
+\beta\mathcal{L}_{\mathrm{KL}},
\label{eq:pgvae_loss}
\end{equation}
where $\lambda_f$, $\lambda_g$, $\lambda_c$ and $\beta$ are hyperparameters. We set $\lambda_f=0.1$, $\lambda_g=1$, $\lambda_c=0.1$, and adapt $\beta$ within $[10^{-6},3\times10^{-3}]$.

\subsection{ViGA: Visual-to-Geometric Adapter}
\label{subsec:enhanced_sapiens2}

We design a Visual-to-Geometric Adapter (ViGA) to predict the corresponding PG-VAE latent vectors from a single image. ViGA is attached to a pretrained visual Transformer, using Sapiens2~\cite{khirodkar2026sapiens2} as the backbone. With the backbone fully frozen, ViGA reads intermediate token features through a side branch without modifying the original parameters, token states, or outputs. This asymmetric design preserves the pretrained semantic representation while enabling ViGA tokens to aggregate contextual visual cues for latent alignment.

\textbf{Model Architecture.}
ViGA maintains learnable token embeddings as model input, comprising four FLAME-semantic and $N_g$ geometric tokens. 
Let $X^l$ denote the native token sequence of the frozen visual encoder at layer $l$. ViGA introduces an extra token sequence $Y^l=[F^l;G^l]$,
where $F^l$ denotes the hidden features of the four FLAME-semantic tokens and $G^l$ denotes those of the geometric tokens.
At each visual-encoder layer, the frozen backbone follows its original forward pass, while ViGA updates only its geometric stream: 

\begin{equation}
X^{l+1}=\mathcal{B}^{l}(X^{l}),\qquad
Y^{l+1}=\operatorname{CA}^{l}\bigl(Y^{l},[X^{l};Y^{l}]\bigr),
\end{equation}
where $\mathcal{B}^{l}$ denotes the frozen backbone self-attention block and $\operatorname{CA}^{l}$ denotes cross-attention whose second argument provides both keys and values. Thus, the geometric tokens can read contextual visual features and interact with one another, but the native visual tokens never attend to or are updated by the geometric tokens. This asymmetric design preserves the original visual representation while adding a dedicated pathway for geometric reasoning.
At the final layer $L_d$, a projection head maps the ViGA token features $[F^{L_d};G^{L_d}]$ to the PG-VAE latent space, yielding $\hat Z=W_z[F^{L_d};G^{L_d}]$.

\textbf{Training Objective.}
The supervision target is the posterior mean produced by the frozen PG-VAE encoder from the canonical target surface, $Z^\star=\operatorname{sg}[\mu_{\mathrm{PG\text{-}VAE}}(V_{\mathrm{gt}})]$. We use a group-balanced alignment objective,
$\mathcal{L}_{\mathrm{align}}
=
\tfrac{1}{2}\bigl(
\|\hat Z_{\mathrm{sem}}-Z^\star_{\mathrm{sem}}\|_2^2
+
\|\hat Z_{\mathrm{geo}}-Z^\star_{\mathrm{geo}}\|_2^2
\bigr)$, 
preventing the more numerous geometric tokens from dominating the four semantic tokens.
During training, only the geometric queries, cross-attention blocks, and latent projection head are optimized. At inference, the frozen PG-VAE decoder maps $\hat Z$ to an ordered geometric reconstruction or interpretable FLAME parameters.

\begin{table*}[htbp]
\centering
\scriptsize
\renewcommand{\arraystretch}{0.95}
\caption{\textbf{Face-only reconstruction on synthetic FLAME subsets and the RenderMe360~\cite{pan2023renderme} dataset.}
Results are averaged over all five camera views after rigid ICP alignment.
CD-L1 is reported in millimeters. Best results are in bold.
$\Delta$ denotes ViGA-FLAME Param. minus the strongest baseline for each metric;
negative CD-L1 and positive F@5 indicate improvements.
}
\label{tab:fixed_yaw_cross_regime}
\begin{tabular}{@{}lrrrrrrrrrrrr@{}}
\toprule
\multirow{2}{*}{Method} &
\multicolumn{2}{c}{All typical} &
\multicolumn{2}{c}{Shape extreme} &
\multicolumn{2}{c}{Expression extreme} &
\multicolumn{2}{c}{Pose extreme} &
\multicolumn{2}{c}{All extreme} &
\multicolumn{2}{c}{RenderMe360~\cite{pan2023renderme}} \\
\cmidrule(lr){2-3}
\cmidrule(lr){4-5}
\cmidrule(lr){6-7}
\cmidrule(lr){8-9}
\cmidrule(lr){10-11}
\cmidrule(lr){12-13}
&
CD-L1$\downarrow$ & F@5$\uparrow$ &
CD-L1$\downarrow$ & F@5$\uparrow$ &
CD-L1$\downarrow$ & F@5$\uparrow$ &
CD-L1$\downarrow$ & F@5$\uparrow$ &
CD-L1$\downarrow$ & F@5$\uparrow$ &
CD-L1$\downarrow$ & F@5$\uparrow$ \\
\midrule
DECA~\cite{feng2021deca}
& 3.181 & 0.864
& 3.468 & 0.822
& 3.285 & 0.849
& 6.460 & 0.650
& 6.642 & 0.608
& 3.972 & 0.747 \\
SMIRK~\cite{filntisis2024smirk}
& 3.267 & 0.848
& 3.489 & 0.819
& 3.341 & 0.835
& 6.016 & 0.653
& 6.260 & 0.618
& 4.184 & 0.723 \\
TEASER~\cite{liu2025teaser}
& 3.443 & 0.821
& 3.662 & 0.792
& 3.517 & 0.812
& 5.620 & 0.642
& 5.817 & 0.615
& 4.492 & 0.693 \\
SHeaP~\cite{schoneveld2025sheap}
& 4.269 & 0.823
& 4.988 & 0.769
& 4.786 & 0.788
& 8.023 & 0.613
& 8.113 & 0.585
& \textbf{3.754} & \textbf{0.780} \\
\midrule
ViGA-FLAME Recon.
& 2.693 & 0.937
& 2.774 & 0.923
& 2.727 & 0.930
& 3.581 & 0.835
& 3.768 & 0.798
& 4.430 & 0.696 \\
ViGA-FLAME Param.
& \textbf{2.556} & \textbf{0.949}
& \textbf{2.710} & \textbf{0.929}
& \textbf{2.659} & \textbf{0.938}
& \textbf{3.212} & \textbf{0.868}
& \textbf{3.445} & \textbf{0.833}
& 4.074 & 0.731 \\
\multicolumn{1}{r}{\emph{$\Delta$ vs.\ best baseline}} &
$-0.625$ & $+0.085$ &
$-0.758$ & $+0.107$ &
$-0.626$ & $+0.089$ &
$-2.408$ & $+0.215$ &
$-2.372$ & $+0.215$ &
$+0.320$ & $-0.049$ \\
\bottomrule
\vspace{-0.6cm}
\end{tabular}
\end{table*}

\begin{table}[t]
\centering
\scriptsize
\setlength{\tabcolsep}{3.0pt}
\renewcommand{\arraystretch}{0.95}
\caption{\textbf{Face-only reconstruction across camera viewpoints.}
Results are averaged across all five synthetic FLAME parameter subsets after rigid ICP alignment.
CD-L1 is reported in millimeters. Best results are in bold.}
\label{tab:cross_viewpoint}
\begin{tabular}{@{}lrrrrrr@{}}
\toprule
\multirow{2}{*}{Method} &
\multicolumn{2}{c}{Frontal ($0^\circ$)} &
\multicolumn{2}{c}{Oblique  ($\pm45^\circ$)} &
\multicolumn{2}{c}{Profile ($\pm90^\circ$)} \\
\cmidrule(lr){2-3}
\cmidrule(lr){4-5}
\cmidrule(lr){6-7}
&
CD-L1$\downarrow$ & F@5$\uparrow$ &
CD-L1$\downarrow$ & F@5$\uparrow$ &
CD-L1$\downarrow$ & F@5$\uparrow$ \\

\midrule
DECA~\cite{feng2021deca}
& 5.263 & 0.756
& 4.387 & 0.775
& 4.499 & 0.744 \\
SMIRK~\cite{filntisis2024smirk}
& 4.394 & 0.779
& 4.373 & 0.774
& 4.616 & 0.723 \\
TEASER~\cite{liu2025teaser}
& 4.116 & 0.772
& 4.407 & 0.745
& 4.564 & 0.710 \\
SHeaP~\cite{schoneveld2025sheap}
& 5.642 & 0.736
& 5.408 & 0.743
& 6.860 & 0.678 \\
\midrule
ViGA-FLAME Recon.
& 3.019 & 0.902
& 3.011 & 0.896
& 3.251 & 0.864 \\
ViGA-FLAME Param.
& \textbf{2.723} & \textbf{0.930}
& \textbf{2.820} & \textbf{0.915}
& \textbf{3.109} & \textbf{0.879} \\
\bottomrule
\vspace{-0.4cm}
\end{tabular}
\end{table}

\subsection{Training Data Synthesis}\label{subsec:data}

We generate training pairs by sampling FLAME parameters and rendering corresponding images with Nvdiffrast~\cite{Laine2020diffrast}, which provides exact correspondence among images, canonical surfaces, and FLAME parameters. We randomize identity, expression, local pose, texture~\cite{bai2023ffhq}, hair~\cite{rosu2025difflocks}, background, and illumination to improve appearance diversity. PG-VAE is trained on FLAME surfaces sampled online, whereas ViGA is trained on $100$K rendered images at $768\times1024$ resolution.

\section{Experiments}
\subsection{Implementation Details}
We initialize PG-VAE attention layers from Hunyuan3D 2.0~\cite{zhao2025hunyuan3d} and optimize newly introduced and transferred parameters using AdamW with learning rates of $10^{-4}$ and $10^{-5}$, respectively, $500$ warm-up steps, a weight decay of $0.01$, and a global batch size of $1024$.  
ViGA is trained on $768\times1024$ images using AdamW with a learning rate of $10^{-4}$, weight decay $0.05$, and a global batch size of $256$. Both stages use bfloat16 mixed-precision training on 4 NVIDIA H200 GPUs.

\subsection{Single-Image FLAME Parameter Estimation}
\begin{figure}[t]
    \centering
    \begin{overpic}[width=\columnwidth]{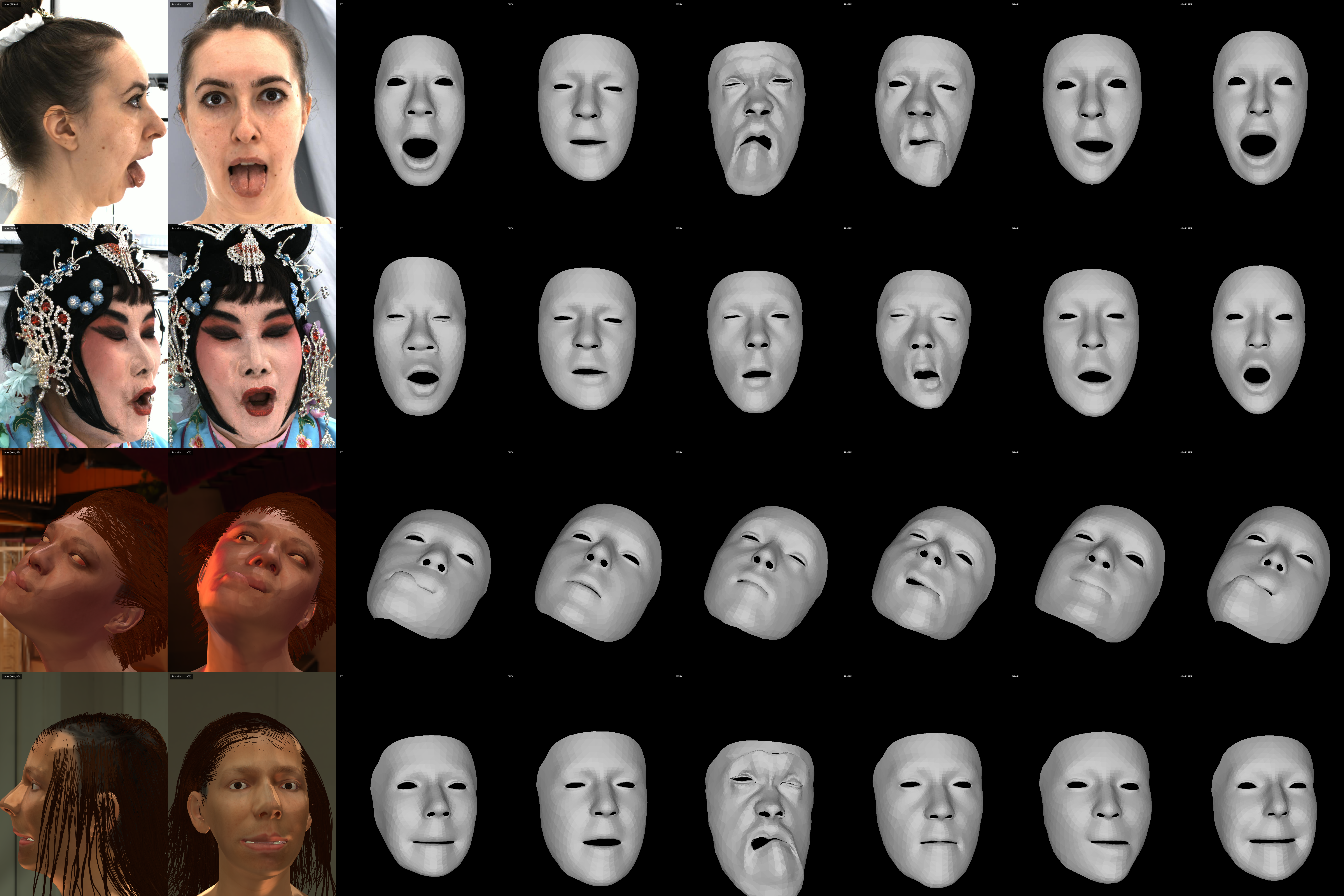}

        \put(0,0){\color{blue}\linethickness{1pt}\framebox(12.5,66.6){}}
        
        \put(87.5,0){\color{red}\linethickness{1pt}\framebox(12.5,66.6){}}
 
        \put(5,-5){\scriptsize\textbf{(a)}}
        \put(17.5,-5){\scriptsize\textbf{(b)}}
        \put(30,-5){\scriptsize\textbf{(c)}}
        \put(42.5,-5){\scriptsize\textbf{(d)}}
        \put(55,-5){\scriptsize\textbf{(e)}}
        \put(67.5,-5){\scriptsize\textbf{(f)}}
        \put(80,-5){\scriptsize\textbf{(g)}}
        \put(92.5,-5){\scriptsize\textbf{(h)}}
    \end{overpic}

    \vspace{0.5mm} 
    \setlength{\abovecaptionskip}{0mm}
    \setlength{\belowcaptionskip}{0mm}
    \caption{\textbf{Qualitative face reconstruction comparisons.}
    Each row shows one example: (a) \textcolor{blue}{Input image}; (b) frontal-view display;
    (c) Scan/GT mesh; (d) DECA~\cite{feng2021deca};
    (e) SMIRK~\cite{filntisis2024smirk}; (f) TEASER~\cite{liu2025teaser};
    (g) SHeaP~\cite{schoneveld2025sheap}; (h) Our \textcolor{red}{ViGA-FLAME}.}
    \vspace{-0.4cm}
    \label{fig:qualitative_comparison}
\end{figure}

We evaluate whether the learned representation generalizes beyond the FLAME2023 geometry used to train our model. We construct a controlled synthetic benchmark with random FLAME parameters. For each identity, we render five fixed camera views corresponding to frontal ($0^\circ$), oblique ($\pm45^\circ$), and profile ($\pm90^\circ$) observations. 
We synthesise head variations by sampling FLAME shape, expression, and pose parameters under five regimes: \emph{all typical}, \emph{shape extreme}, \emph{expression extreme}, \emph{pose extreme}, and \emph{all extreme}. Each regime contains $50$ instances. All factors remain within their typical ranges in \emph{all typical}, while only the specified factor is extreme in each single-factor regime and all factors are extreme in \emph{all extreme}. For shape and expression, 
typical PCA coefficients lie within $2.5\sigma$; extreme shape/expression regimes set at least one leading $20$ coefficient between $2.5\sigma$ and $3.5\sigma$, while extreme pose regimes use rotations near the physical limits of the neck, jaw, and eyes. Real-world evaluations use RenderMe360~\cite{pan2023renderme}.
We compare the reconstruction and parametric outputs of ViGA-FLAME with FLAME-based regressors: DECA~\cite{feng2021deca}, SMIRK~\cite{filntisis2024smirk}, TEASER~\cite{liu2025teaser}, and SHeaP~\cite{schoneveld2025sheap}. We evaluate only a common face-only surface region. Each predicted surface is aligned to the FLAME ground-truth face mesh using rigid ICP with rotation and translation only. We report the averaged symmetric Chamfer-L1 distance (CD-L1, mm) and the F-score at a $5$\,mm threshold (F@5).

As shown in Table~\ref{tab:fixed_yaw_cross_regime}, both ViGA-FLAME outputs outperform all baselines across the synthetic regimes, with the parametric branch achieving the strongest results under the most challenging \emph{all extreme} setting, highlighting robustness when shape, expression, and pose vary jointly.
Table~\ref{tab:cross_viewpoint} further shows that the parametric branch achieves the best reconstruction accuracy at frontal, oblique, and profile viewpoints. 
On the real-world RenderMe360 dataset, our parametric branch surpasses SMIRK and TEASER, though trailing DECA and SHeaP. This transfer is notable as ViGA-FLAME is trained exclusively on synthetic renders, whereas baselines leverage real-image supervision or self-supervision. The residual gap reflects limited synthetic appearance diversity (e.g., heavy makeup, age-related skin textures), pointing to data augmentation rather than a flaw in our latent-alignment formulation.
Moreover, optimization-based methods rely on 2D landmark fitting, whose localization becomes unreliable under oblique and profile views due to severe self-occlusion. We therefore restrict quantitative comparisons to feed-forward methods operating under a matched single-image inference protocol.

\subsection{Ablation Studies}
\noindent\textbf{PG-VAE.}
We evaluate five variants on 8k randomly sampled FLAME parameters. We report mean corresponding-vertex error distance in millimeters for the topology-aware reconstruction branch (Recon.), the parametric FLAME branch (Param), and the cross-branch discrepancy (Recon.$\leftrightarrow$Param.). 
Lower values indicate more accurate geometry or stronger agreement between the two decoding pathways.

\begin{table}[htbp]\centering\small
\vspace{-0.2cm}
\setlength{\tabcolsep}{3pt}
\caption{Ablation studies of \headvae~variations.} \label{tab:ablation_headvae}
\begin{tabular}{lccc}\toprule
Method & Recon. ↓ & Param. ↓ & Recon.$\leftrightarrow$Param. ↓ \\ \midrule
Base & 1.0281 & 0.9472 & 1.2707\\
+ $\lambda_g\mathcal{L}_{\mathrm{par\text{-}gt}}$ & 0.9942 & 0.8152 & 1.1726\\
++ $\lambda_c\mathcal{L}_{\mathrm{par\text{-}geo}}$ & 1.0347 & 0.7643 & 1.2344\\
+++ $\text{Cond}_{query}$ & 0.8647 & 0.4975 & 0.9713\\ 
+++ $\text{Cond}_{add}$ & 0.9831 & 0.5830 & 1.1605\\ \bottomrule
\vspace{-0.2cm}
\end{tabular}\end{table} 

Table~\ref{tab:ablation_headvae} shows that parametric-surface supervision $\mathcal{L}_{\mathrm{par\text{-}gt}}$ improves all metrics over the baseline, 
confirming that direct supervision in canonical vertex space complements coefficient-level FLAME regression. Adding the cross-branch consistency term, $\mathcal{L}_{\mathrm{par\text{-}geo}}$, further improves parametric reconstruction, despite slightly degrades the geometric reconstruction and measured branch discrepancy. This trade-off suggests that consistency gradients can perturb the geometric branch before the parametric branch converges.

We compare unconditional learnable semantic embeddings with parameter-conditioned semantic tokens. Replacing semantic query embeddings with factor-specific MLP projections of FLAME factors achieves the best performance on all metrics, outperforming additive fusion. 
Thus, factor-specific conditioning provides a stronger inductive bias for organizing semantic token slots and aligning parametric decoding with dense geometric reconstruction.

\section{Conclusion}
We present ViGA-FLAME, a feed-forward framework aligning a frozen image encoder with PG-VAE's joint parametric-geometric latent space, enabling dense surface reconstruction and FLAME estimation from a single image. Experiments demonstrate improved robustness under controlled variations in shape, expression, pose, and viewpoint, while showing transfer to real-world head estimation.

\section{Acknowledgements}
This research was supported by the Theme-based Research Scheme, Research Grants Council of Hong Kong (T45-205/21-N), and the Guangdong and Hong Kong Universities “1+1+1” Joint Research Collaboration Scheme (2025-A0505000003).

\bibliographystyle{IEEEbib}
\bibliography{myRefs}

\end{document}